\documentclass[letterpaper, 10 pt, journal, twoside]{IEEEtran}
\usepackage{graphics}           
\usepackage{times}              
\usepackage{amsmath}            
\usepackage{amssymb}            
\usepackage{graphicx}
\usepackage{algorithm}
\usepackage[noend]{algpseudocode}
\usepackage{booktabs}
\usepackage{color}
\definecolor{instructioncolor}{rgb}{.5,.5,.5}

\usepackage[font=small]{caption}

\def\secref#1{Sec.~\ref{#1}}
\def\figref#1{Fig.~\ref{#1}}
\def\tabref#1{Tab.~\ref{#1}}
\def\eqref#1{Eq.~(\ref{#1})}

\makeatletter
\usepackage{xspace}
\DeclareRobustCommand\onedot{\futurelet\@let@token\@onedot}
\def\@onedot{\ifx\@let@token.\else.\null\fi\xspace}

\def\etal{{et al}\onedot}
\makeatother

\def\etalcite#1{\etal~\cite{#1}}

\usepackage{array}
\newcolumntype{L}[1]{>{\raggedright\let\newline\\\arraybackslash\hspace{0pt}}m{#1}}
\newcolumntype{C}[1]{>{\centering\let\newline\\\arraybackslash\hspace{0pt}}m{#1}}
\newcolumntype{R}[1]{>{\raggedleft\let\newline\\\arraybackslash\hspace{0pt}}m{#1}}

\usepackage{subfigure}
\usepackage{url}
\usepackage{multirow}
\usepackage{algorithm}
\usepackage{algpseudocode}
\usepackage{amsmath}

\newcommand\chen[1]{\textcolor{black}{#1}}
\newcommand\revise[1]{\textcolor{black}{#1}}

\title{CCL: Continual Contrastive Learning \\ for LiDAR Place Recognition}

\author{Jiafeng Cui,  Xieyuanli Chen$^*$% <-this % stops a space
  \thanks{Manuscript received: Dec. 24, 2022; Revised: Feb. 28, 2023; Accepted: May 20, 2023. This paper was recommended for publication by
  Editor Sven Behnke upon evaluation of the Associate Editor and Reviewers' comments.}
  \thanks{Both authors are with the College of Intelligence Science and Technology, National University of Defense Technology, China. This work has partially been funded by the National Science Foundation of China under Grant U1913202, U22A2059, and 62203460, Major Project of Natural Science Foundation of Hunan Province (No. 2021JC0004)}%
  \thanks{*Corresponding author: Xieyuanli Chen, \protect\url{xieyuanli.chen@nudt.edu.cn}}
}

\begin{document}

\maketitle

\markboth{IEEE Robotics and Automation Letters.}
{J. Cui and X. Chen: CCL: Continual Contrastive Learning for LiDAR Place Recognition}

%%%%%%%%%%%%%%%%%%%%%%%%%%%%%%%%%%%%%%%%%%%%%%%%%%%%%%%%%%%%%%%%%%%%%%%%%%%%%%%%
\begin{abstract}
Place recognition is an essential and challenging task in loop closing and global localization for robotics and autonomous driving applications.
Benefiting from the recent advances in deep learning techniques, the performance of LiDAR place recognition (LPR) has been greatly improved.
However, current deep learning-based methods suffer from two major problems: poor generalization ability and catastrophic forgetting.
In this paper, we propose a continual contrastive learning method, named CCL, to tackle the catastrophic forgetting problem and generally improve the robustness of LPR approaches. 
Our CCL constructs a contrastive feature pool and utilizes contrastive loss to train more transferable representations of places.
When transferred into new environments, our CCL continuously reviews the contrastive memory bank and applies a distribution-based knowledge distillation to maintain the retrieval ability of the past data while continually learning to recognize new places from the new data.
\revise{We thoroughly evaluate our approach on Oxford, MulRan, and PNV datasets using three different LPR methods.
The experimental results show that our CCL consistently improves the performance of different methods in different environments outperforming the state-of-the-art continual learning \chen{approaches}.
% To the best of our knowledge, this is the first work effectively applying continual contrastive learning for LiDAR place recognition. 
The implementation of our method has been released at \protect\url{https://github.com/cloudcjf/CCL}.}
\end{abstract}


%%%%%%%%%%%%%%%%%%%%%%%%%%%%%%%%%%%%%%%%%%%%%%%%%%%%%%%%%%%%%%%%%%%%%%%%%%%%%%%%
\section{Introduction}
\label{sec:intro}

Place recognition is a key component of simultaneous localization and mapping (SLAM) and global localization for autonomous mobile systems~\cite{durrant2006simultaneous}. 
It identifies previously visited places in GPS-denied environments and provides a rough global location of the robot within the map or database.
LiDAR-based place recognition (LPR) is more robust against illumination and viewpoint changes than its camera-based counterparts since LiDAR sensors provide geometric structural information of the environment with a wide horizontal view~\cite{cui2022dsc}.
Benefiting from modern deep learning techniques, many LPR methods have been proposed \chen{\cite{angelina2018cvpr,liu2019lpd,chen2021auro,vidanapathirana2022logg3d,komorowski2021minkloc3d,ma2022ral,xu2021disco}} and achieve good place recognition performance in outdoor large-scale environments. 

Despite showing impressive results in recent years, the learning-based LPR methods based on pre-trained networks may easily collapse in unseen environments~\cite{li2021generalizing}.
This is because most existing LPR methods exploiting supervised metric-learning techniques only optimize the network models upon the training data. 
Thus the learned representations usually overfit the training distribution while failing to be transferred to new domains when applied to previously unseen environments.
Fine-tuning~\cite{tajbakhsh2016itmi} the model with the data from the new environment can avoid such degradation. 
However, the expertise learned from the previous domain may distort during the fine-tuning, resulting in bad performance when applied back to former environments~\cite{kumar2022iclr}. 
% While such repeatedly applied in different places is a frequent case for place recognition in robotics and autonomous driving.
This dilemma is well-known as catastrophic forgetting~\cite{kirkpatrick2017overcoming} in deep learning and a bottleneck of the current metric learning-based neural networks for LPR.

\begin{figure}[t]
  \centering
  \includegraphics[width=\linewidth]{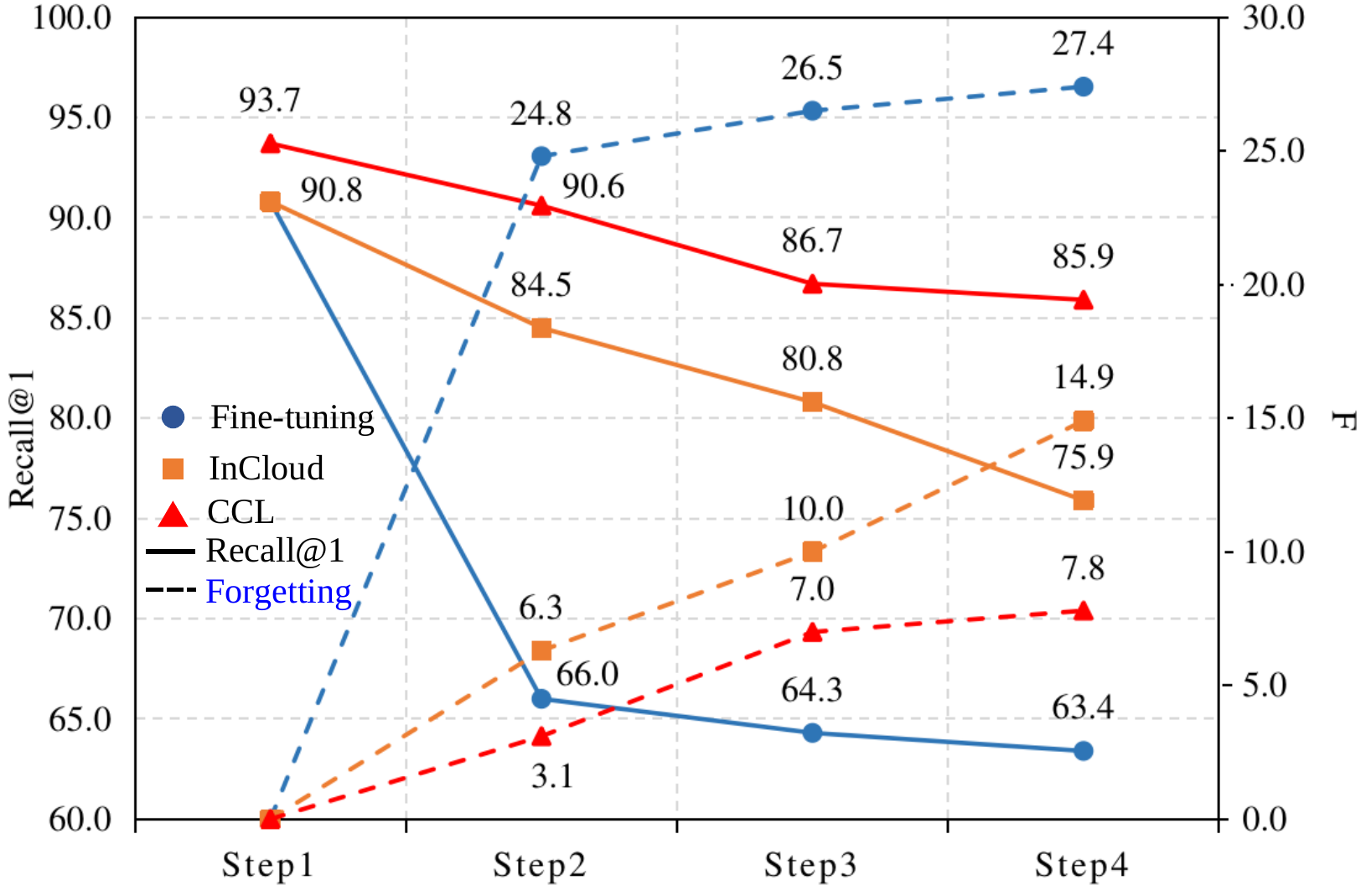}
  \caption{Comparison of different methods in dealing with catastrophic forgetting for LPR. We train the MinkLoc3D model on the Oxford \revise{training dataset in the first step and then continually train it on three new datasets in the following steps. In each step, we evaluate the trained model on the Oxford evaluation dataset and show the Recall@1 and F scores of different continual learning methods. Solid lines represent the Recall@1 metric, while dashed lines indicate the F-score. A higher Recall@1 value indicates superior place recognition results, while a lower F-score suggests less forgetting. As depicted in the figure, vanilla Fine-tuning (blue line) shows the most pronounced reduction in Recall@1 and exhibits the highest degree of forgetting after training on four datasets. 
  InCloud (orange line) demonstrates the ability to mitigate forgetting, albeit with a significant decrease in performance. 
  Our proposed method, CCL (red line), is more effective in addressing catastrophic forgetting and achieves superior performance compared to the other methods.
  }
}
  \label{fig:motivation}
  \vspace{-0.4cm}
\end{figure}

To tackle the \chen{aforementioned} problem, continual learning~\cite{li2017tpami} has been introduced to place recognition in previous work such as AirLoop~\cite{gao2022airloop} and InCloud~\cite{knights2022incloud}.
They utilize a small set of past samples with current data to train the network, enabling the model to learn \chen{representations} of new environments while preserving past knowledge. 
\revise{Despite the effectiveness of these methods in mitigating catastrophic forgetting in LPR, their reliance on triplet-based metric learning can pose limitations when applied to unseen environments. 
Triplet-based metric learning, which updates modeling based on limited information from triplets, can introduce biases that hinder the transferability of the learned representations in place recognition tasks.
However, the forgetting observed in the place recognition task is not solely caused by the distortion of experience but also by less transferable representations lacking utility in long-term missions.
Therefore, more transferable descriptors are needed to better enable continual learning reducing catastrophic forgetting for place recognition.}

\revise{The main contribution of this paper is a novel learning scheme that combines contrastive and continual learning, dubbed CCL, which extracts more transferable descriptors to tackle the catastrophic forgetting problem and improves the generalization ability of LPR methods, as shown in~\figref{fig:motivation}.}
More specifically, we train LPR methods in a contrastive way that exploits a large contrastive feature pool and optimizes the network model using contrastive loss, thus extracting more descriptive but generalized features from the LiDAR data.
Based on such transferable features, our CCL then conducts continual training.
For each LiDAR scan, we construct a training unit consisting of its corresponding positive and negative pairs. We compose the feature pool using the past data from the memory bank and current data and treat all the past data as the negative pairs of the current data and vice versa.
In the end, our CCL applies a feature distribution-based knowledge distillation from past samples between adjacent training steps to further overcome catastrophic forgetting.  
We thoroughly evaluate our method on three LPR methods with multiple datasets collected by different LiDAR sensors in various environments. The experimental results show our approach significantly outperforms the state-of-the-art continual learning method in overcoming catastrophic forgetting and consistently improves LPR methods' generalization ability. 

\revise{
To the best of our knowledge, our work is the first to address catastrophic forgetting in LPR using continual contrastive learning. We make the following three claims that our proposed CCL method:
(i) overcomes catastrophic forgetting with a novel continual learning strategy and outperforms the state-of-the-art continual learning method in dealing with catastrophic forgetting for LPR methods;
(ii) achieves stronger generalization ability across different datasets compared to existing metric learning schemes;
(iii) can be applied to different place recognition networks, and used as a common strategy to improve the robustness of long-term LPR methods.
These claims are supported by extensive experiments and ablation studies conducted in this study.
}
% To sum up, the main contributions of this paper are:  
% (i) Our proposed CCL is the first work introducing continual contrastive learning for LPR. It outperforms the state-of-the-art continual learning method in dealing with catastrophic forgetting for LPR methods;
% %
% (ii) Our CCL enables LPR methods to achieve stronger generalization ability compared to existing metric learning schemes;
% %
% (iii) Extensive experiments and ablation studies using different LiDAR place recognition methods show that our approach generally and significantly improves LPR performance, which can be used as a common strategy to improve the robustness of long-term LPR methods. 

\begin{figure*}[ht] 
        \centering
        \includegraphics[width = 0.9\textwidth]{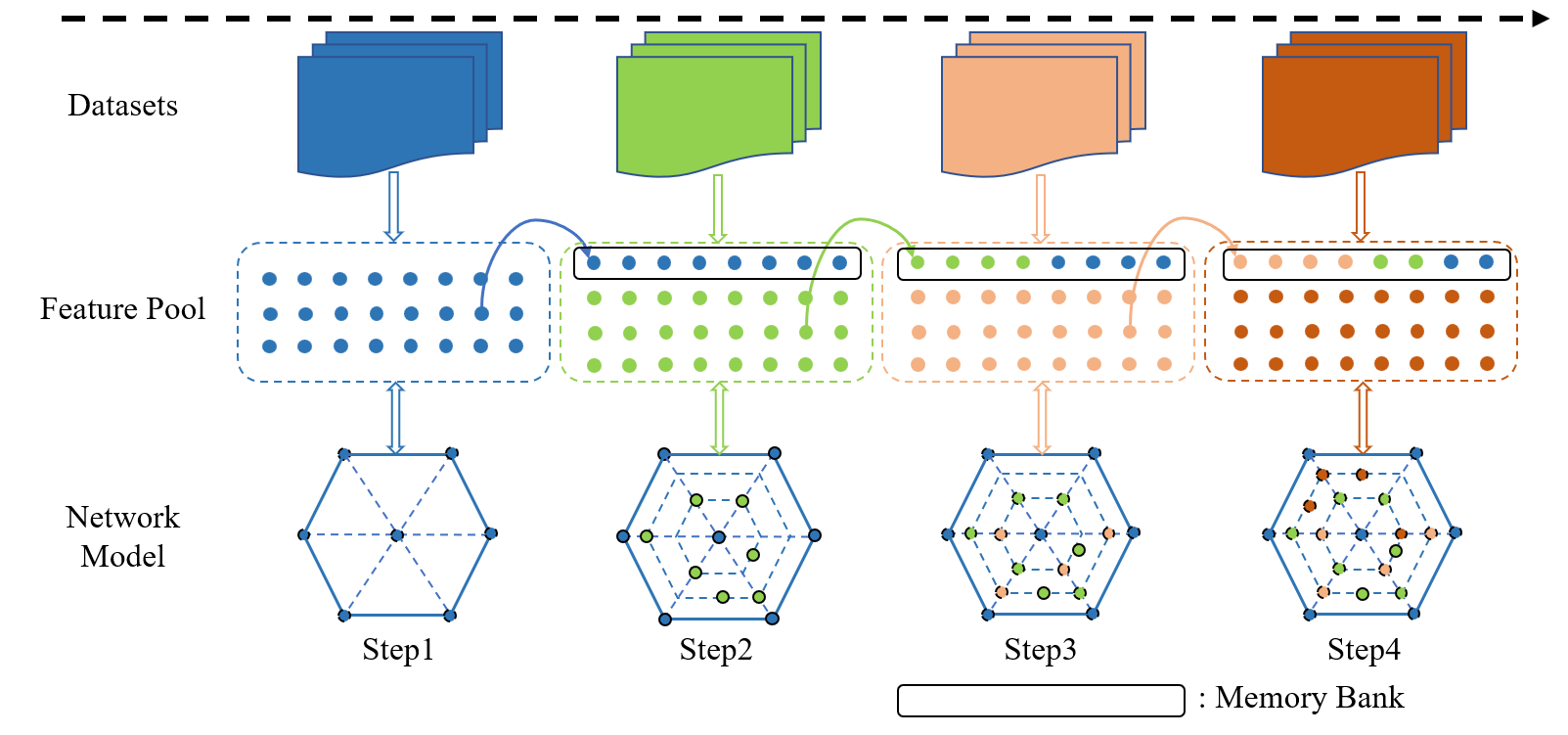} 
        \caption{
        The pipeline of our CCL. Datasets are coming and fed into our model by sequence.
        The feature pool contains a large number of representations. 
        It consists of two parts, feature representations of the current dataset and the memory bank review of past datasets.
        On the bottom, the model is trained in a contrastive way in each step.
        After updating the model parameters, the current mini-batch of representations will be added to the feature pool and replace the corresponding early representations.
        With continual training, the model learns various representations from all the environments while well preserving the structure.}
        \label{fig:overview}
        \vspace{-0.4cm}
\end{figure*}

%%%%%%%%%%%%%%%%%%%%%%%%%%%%%%%%%%%%%%%%%%%%%%%%%%%%%%%%%%%%%%%%%%%%%%%%%%%%%%%%
\section{Related Work}
\label{sec:related}

Place recognition is a classic topic with many published works in computer vision, robotics, and autonomous driving. Here, we mainly focus on methods using LiDAR sensors, especially the recent learning-based methods.

There are two large groups of LPR methods, the projected image-based methods~\chen{\cite{chen2021auro,ma2022ral,ma2022tie,xu2021disco}} and point cloud-based methods~\cite{angelina2018cvpr,liu2019lpd,komorowski2021minkloc3d,vidanapathirana2022logg3d}. 
Using projected image-based representations, networks can easily deal with sparse LiDAR data in traditional image-based fashion and achieve fast operation that can be deployed for online real-world applications. 
However, such projection methods may only work for specific rotating LiDAR sensors, thus less transferable to new sensors applied in new environments. 
In contrast, point cloud-based methods directly operate on individual points or voxels and generalize better to new sensors. 
\revise{PointNetVLAD proposed by Uy~\etalcite{angelina2018cvpr} is a pioneering point-based network for LPR.}
It uses PointNet~\cite{qi2017pointnet} to extract point-wise features and feed them to NetVLAD~\cite{arandjelovic2016netvlad} to generate the global descriptor of each LiDAR scan for place recognition.
% The structure of the network is simple and small to train, but the recognition performance is far from satisfaction.
Based on PointNetVLAD, LPD-Net by Liu~\etalcite{liu2019lpd} uses DGCNN~\cite{wang2019dynamic} as a point-wise feature extractor, which considers the neighborhood information of each point and improves the LPR performance.
Unlike previous networks, MinkLoc3D by Komorowski~\etalcite{komorowski2021minkloc3d} transforms the unordered point cloud data to 3D voxels where classic convolution \chen{operations} can be applied.
It introduces a sparse convolutional backbone~\cite{choy20194d} and a more lightweight GeM~\cite{radenovic2018fine} pooling to generate a global descriptor for LPR, which speeds up the computationally intensive convolution operation.
LoGG3D-Net by Vidanapathirana~\etalcite{vidanapathirana2022logg3d} also uses a sparse convolutional backbone and provides a coarse pose estimation between point clouds.
Most recently, Ma~\etalcite{ma2023arxiv} exploit cross attention mechanism to fuse different views of LiDAR data and improve the LPR performance.
Although these methods have shown impressive performance in the trained environments, they cannot generalize well in unseen environments based on metric learning.

Contrastive learning can generate more general representations and has been widely studied in computer vision tasks, such as visual representation and image classification~\cite{khosla2020arxiv,chen2020icml,he2020cvpr-mcfu}.
% It is capable of learning transferable features without labeling.
Gadd~\etalcite{gadd2021contrastive} first introduces contrastive learning to radar place recognition and achieves considerable performance on various radar datasets. 
% The positive and negative pairs are divided using the timestamps of frames.
Most recently, KPPR by Wiesmann~\etalcite{wiesmann2022kppr} also exploits contrastive learning to improve LPR performance. 
Both use contrastive learning to generate more descriptive features for place recognition, while the catastrophic forgetting problem is not considered.
Fine-tuning the pre-trained model is needed for existing methods to generalize into new environments, which distorts the pre-trained models and causes catastrophic forgetting~\cite{kumar2022iclr}.  

Continual learning~\cite{li2017tpami} has recently been introduced to tackle the catastrophic forgetting problem. The typical continual learning configuration assumes that training data comes sequentially, and models should learn from new data while maintaining the knowledge from the past.
Such a setup fits well the robotics and autonomous driving applications, where the vehicle travels in different environments and learns online.
However, few works have been done on using continual learning to alleviate catastrophic forgetting of place recognition.
AirLoop by Gao~\etalcite{gao2022airloop} first pays attention to forgetting in visual place recognition tasks. 
Inspired by that, InCloud by Knights~\cite{knights2022incloud} introduces continual learning to LPR very recently.
They utilize metric learning with triplet loss for every step in the training process, where a set of past samples is exploited together with current training data so that the model can review the past knowledge.
However, forgetting comes from both limited access to experience and less transferable features that are not useful for long-term tasks.

In contrast to the above works, we are the first to combine contrastive and continual learning in place recognition, addressing the poor generalization and catastrophic forgetting problems simultaneously for LPR methods.

%%%%%%%%%%%%%%%%%%%%%%%%%%%%%%%%%%%%%%%%%%%%%%%%%%%%%%%%%%%%%%%%%%%%%%%%%%%%%%%%
\section{Continual Contrastive Learning for \\ LiDAR Place Recognition}
\label{sec:main}

This work aims to tackle the continual LPR task (see~\secref{sec:lpr-setup}). 
To this end, we propose a novel learning scheme, named CCL, with a special focus on overcoming catastrophic forgetting and improving the generalization ability of LPR methods. 
As shown in~\figref{fig:overview}, our CCL combines contrastive and continual learning naturally and effectively.
It builds a contrastive feature pool and applies contrastive loss to improve the generalization of the descriptors generated by LPR methods (see~\secref{sec:contrastive}). During the continual place recognition, it maintains a memory bank using the contrastive features to review the previous knowledge (see \secref{sec:mr}) and applies a feature distribution-based knowledge distillation (see~\secref{sec:kd}) to overcome catastrophic forgetting.

%%%%%%%%%%%%%%%%%%%%%%%%%
\subsection{Continual LiDAR Place Recognition Setup}
\label{sec:lpr-setup}

We first define the continual LPR task.
Given an input LiDAR scan $\mathcal{P}$, most LPR approaches first encode the input scan into a D-dimension descriptor $f \in \mathbb{R}^{D}$.
Network training aims to reduce the distances in the embedding space between two descriptors of two geographically close scans while enlarging the feature distances if two scans are far away.
Specifically, for any point clouds $\{\mathcal{P}_{i}, \mathcal{P}_{j}, \mathcal{P}_{k}\}$, if
$d(\mathcal{P}_{i}, \mathcal{P}_{j}) < d(\mathcal{P}_{i}, \mathcal{P}_{k})$, then $\left \| f_{i} - f_{j} \right \| _2 < \left \| f_{i} - f_{k} \right \| _2$ where $d(\cdot)$ means the geographic distance between two point clouds and $\left\|\cdot\right\|_2$ is the $L_2$ distance between the corresponding features.

In real-world applications, the LPR networks should be capable of recognizing places in different environments. 
Suppose there are a couple of datasets $\textstyle \bigcup _{t=1}^{T}\mathcal{S}_{t}$, and each dataset $\mathcal{S}_{t}$ covering one environment has a different domain from the others. In that case, the continual learning setup assumes that the robot obtains the datasets by sequence.
Therefore, the historical data may become unavailable due to the limited onboard memory when the robot travels in new environments.
LPR methods need to be trained on sequential datasets continually to learn from the current data and adapt to the new domain. 
\revise{Thus, the challenge of achieving continual place recognition resides in guaranteeing that the updated networks, which have been further trained using data from new environments, uphold their performance when deployed in past locations.}

%%%%%%%%%%%%%%%%%%%%%%%%%
\subsection{Asymmetric Contrastive Learning}
\label{sec:contrastive}

To achieve long-term and continual place recognition, the LPR methods need to train more transferrable network models. 
In this paper, we propose to use contrastive learning for LPR.
% We now illustrate the Asymmetric Supervised Contrastive Loss in detail. 
The original contrastive learning~\cite{oord2018representation} trains a network in an unsupervised fashion with the loss function as:
\begin{align}
    L_{c} = -\log\frac{\exp(f_q\cdot {f^\top_{q'}}/\tau )}{ {\textstyle \exp(f_q\cdot {f^\top_{q'}}/\tau ) + \sum_{i=1}^{K}\exp(f_q\cdot f^\top_{n_i}/\tau)} }
\end{align}
where $f_q$ represents the feature of a query sample, $f_{q'}$ is the feature of positive sample augmented from query, and $\{f_{n_{i}}\}_{i=1}^K$ are features of $K$ negatives from all other samples besides the query.
$\tau$ is a temperature hyperparameter to weigh negative samples.
\revise{The loss aims to enlarge the similarity between the query and the positive and the dissimilarity between the query against all the negatives.}

\revise{The current contrastive learning methodology was developed specifically for image classification tasks and cannot be directly applied to the LPR task. This is because only considering the augmented point clouds as positive samples for a query scan is inadequate, as neighboring scans with different viewpoints can also be obtained from the same location.
To exploit contrastive learning for LPR, we adopt positive and negative samples based on geographic distance~\cite{angelina2018cvpr}. 
Specifically, the training data is represented as $\{q,\{p_{j}\},\{n_{i}\}\}$, where $p_{j}$ and $n_{i}$ represent positive and negative samples of a query sample $q$.}

\begin{figure}[t]
  \centering
  \includegraphics[width=0.95\linewidth]{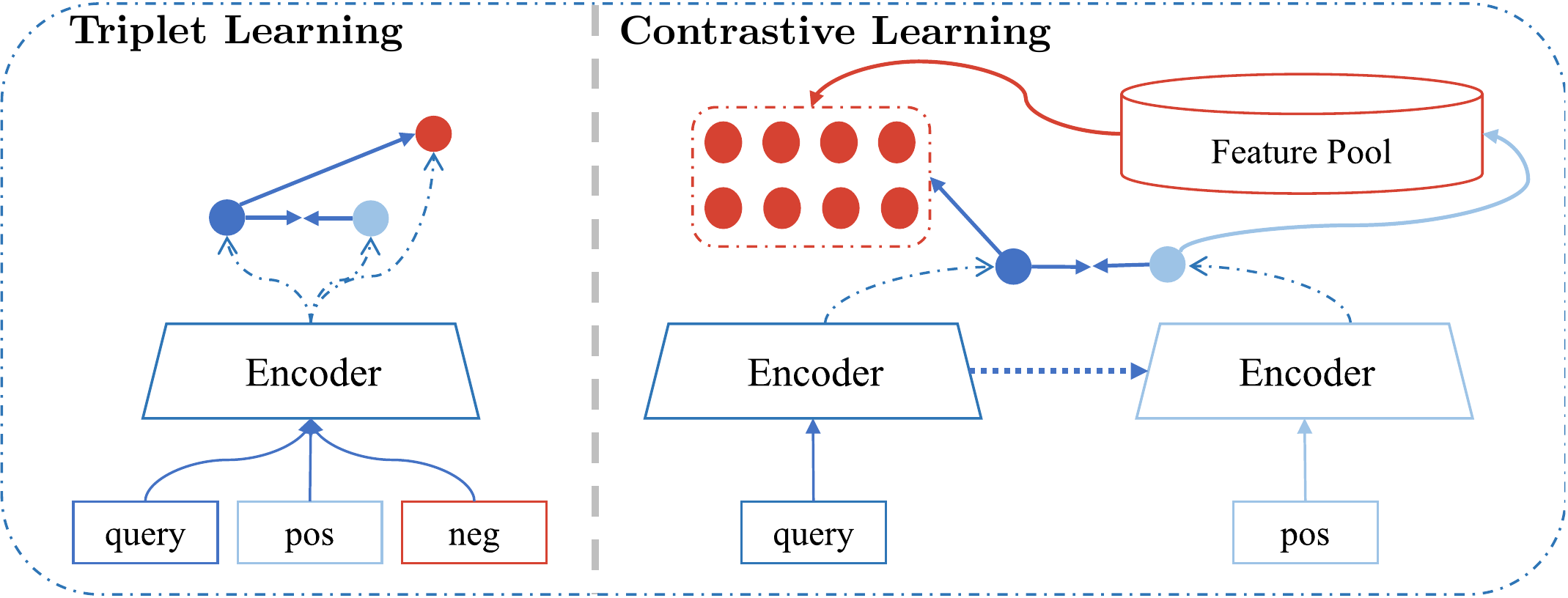}
  \caption{Comparison between triplet  and contrastive learning.
  The triplet-based method only learns from one pair of samples each time, while the contrastive-based method learns from thousands of negative samples, and more informative representations can be learned.}
  \label{fig:tri_con}
  \vspace{-0.3cm}
\end{figure}

Based on the contrastive samples, we slightly adapt LPR network architectures to enable contrastive learning, as shown in~\figref{fig:tri_con}.
We use two independent encoders, a query encoder extracting query features and a positive encoder extracting positive features. 
We utilize a feature pool to dynamically update the negative features from the last batch of the positives.
\revise{To maintain the consistency in feature pool and improve the training efficiency, a momentum update is used for the positive encoder~\cite{he2020cvpr-mcfu}:
\begin{align}
    \Theta^k_{pos} = m\Theta^{k\text{-}1}_{pos} + (1 - m)\Theta^k_{query}
\end{align}
where the positive encoder $\Theta^k_{pos}$ after $k$\,th batch is updated by the $k\text{-}1$\,th positive encoder and the $k$\,th query encoder $\Theta^k_{query}$ with a momentum $m$.
After each inference, we store positive features in the feature pool serving as potential negatives for future batches.}

We furthermore add a small projection head to LPR backbones after the final descriptor when applying the contrastive loss to LPR networks during the training.
\revise{Since the final output is trained to be invariant to data augmentation by contrastive loss~\cite{chen2020icml}, it may ignore useful information of place recognition, such as objects' orientation.} Thus, leveraging the projector can maintain more information in the original global descriptor.
We just use a nonlinear projection head $g(\cdot)$ after the output to keep our CCL simple and can be easily applied to other LPR methods. It consists of an MLP with one hidden layer to obtain a projected feature $z = g(f) = W^{(2)}\sigma (W^{(1)}\cdot f)$, where $\sigma$ is a ReLU operator. We then apply the asymmetric contrastive loss on projected features:
\begin{align}
    L_{c} = -\log\frac{\exp(z_q\cdot z^\top_{p_{j}}/\tau )}{ {\textstyle \exp(z_q\cdot z^\top_{p_{j}}/\tau ) + \sum_{i=1}^{K}\exp(z_q\cdot z^\top_{n_{i}}/\tau)} }
\end{align}

During the inferring, the small projection head will be dropped, and the original descriptor $f$ will be used to represent the place. 
We will discuss the effectiveness of the projection head detailed in the experiments section.

%%%%%%%%%%%%%%%%%%%%%%%%%
\subsection{Continuously Memory Review}
\label{sec:mr}

To overcome catastrophic forgetting, we propose a natural continuous memory review.
In line with existing continual learning setups~\cite{gao2022airloop, knights2022incloud}, we also assume only a small part of training samples from the previous environment can be maintained due to the limited storage.
The key challenge here is how to efficiently use the limited past samples to help the network recall the memory of the past while learning in the new environment.
Unlike the existing methods using the past and current data separately, our CCL fuses them using the contrastive memory bank to better distinguish the places and boost the LPR performance.

Specifically, following~\cite{gao2022airloop, knights2022incloud}, in the training step $t$, we randomly select $h$ samples along with one of their positives from past datasets from step $1$ to step $t-1$ and construct a $2h$ sized memory bank $\mathcal{M}_t$. The oldest memory samples will be replaced with the newest memory samples using a sliding window.
Besides that, we propose to use more contrastive negative samples by fusing the past and current data.
Since only a small amount of past samples is preserved, most negative pairs cannot be obtained for contrastive learning.
To solve this problem, we treat all the current samples $S_t$ as the negatives of the memory samples $\mathcal{M}_t$ and vice versa. 
This makes the model learn not only from the current domain but also from the past. 
Note that we do not require extra samples from the past and only freely enlarge the contrastive feature pool with more negative samples for free from different datasets.

Our memory review is elegant in two aspects: (i)~Firstly, since the samples from two datasets are obtained from different places under the continual place recognition assumption, and thus we can directly treat them as negative samples without any extra processing. (ii)~Furthermore, it combines contrastive and continual learning naturally in each step to boost the generalization while simultaneously reviewing the memory to deal with catastrophic forgetting.

%%%%%%%%%%%%%%%%%%%%%%%%%
\begin{figure}[t]
  \centering
  \includegraphics[width=0.85\linewidth]{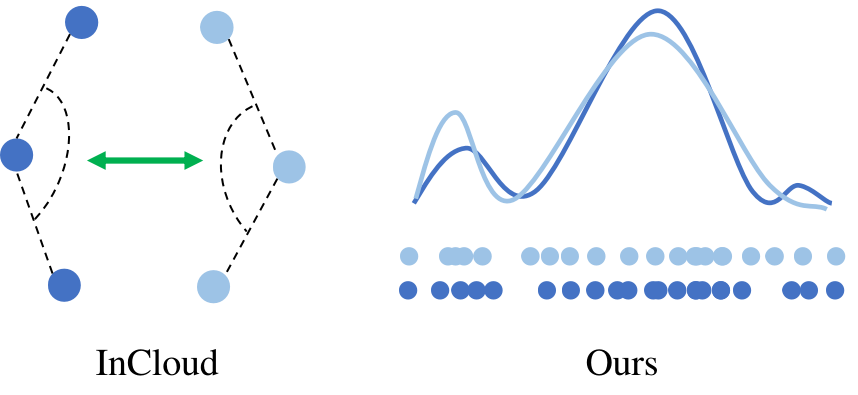}
  \caption{Comparison between the angular-based distillation by InCloud and the proposed distribution-based distillation by our CCL. Distilling the angular relationship only preserves the local structure of features in embedding space, while ours focus more on the global distribution.}
  \label{fig:kd}
  \vspace{-0.3cm}
\end{figure}

\subsection{Knowledge Distillation using Continual Learning}
\label{sec:kd}

% Although contrastive learning generates a generalized representation, it is necessary to introduce a distillation loss to explicitly preserve the learned knowledge. 
\revise{Based on the continuously reviewed memory bank, we then conduct knowledge distillation to explicitly preserve the learned knowledge. 
Unlike existing works~\cite{gao2022airloop, knights2022incloud} using all samples to regularize model parameters, we argue that the model from $t-1$ step has not been trained on the current environment and only memory samples $\mathcal{M}_t$ should be used for distillation. 
Inspired by~\cite{ferdinand2022attenuating}, we propose a feature distribution-based knowledge distillation for LPR shown in~\figref{fig:kd}. Specifically, we infer the contrastive features $z^t_i$ and $z^{t-1}_i$ for each memory sample using the current model and the last frozen model parameters.
We compute the pairwise similarities between $z^t_i$ and $z^t_j$ for the current model, $z^{t-1}_i$ and $z^{t-1}_j$ for the last frozen model and group them into matrices $B^t$ and $B^{t-1}$:
\begin{align}
    B^t_{i,j} &= \frac{z^t_i \cdot (z^{t}_j)^\top}{\tau'} \\
    B^{t-1}_{i,j} &= \frac{z^{t-1}_i \cdot (z^{t-1}_j)^\top}{\tau'}
\end{align}
where $\tau'$ is a distillation temperature hyperparameter.
We calculate the probability distributions of $B^t$ and $B^{t-1}$ and minimize the divergence between those two probability matrices. 
The distillation loss can be computed as:
\begin{align}
    L_{\text{KD}} = \sum_{i,j} \text{KL}\{\text{softmax}(B^{t-1}_{i,j}) \log{(\text{softmax}(B^t_{i,j}))}\}
\end{align}
where softmax is applied for each row.}

\revise{
The total loss of our method CCL is finally defined as:
\begin{align}
    L = L_c + \lambda\,L_{\text{KD}}
\end{align}
where $\lambda$ is determined according to the ratio of memory bank size against the current data size.}
% \begin{align}
%     \lambda = \frac{N_{\mathcal{M}_t}}{N_{S_t}}
% \end{align}

%%%%%%%%%%%%%%%%%%%%%%%%%
% \subsection{Algorithm Details}
% \label{sec:ad}
% Here, we give a complete procedure of our method and additional details. The full algorithm is provided in Algorithm 1\ref{alg:CCL}.
% \todo{We don't need pseudocode but more description of the algorithm}
% \begin{algorithm}[h]
%   \caption{CCL}
%   \label{alg:CCL}
%   \begin{algorithmic}[1]
%     \Require
%       Real-time task set $S_H$

%     \State $D=0$;

%     \For {${\tau _i}:{S_H}$}
%         \State $D = D+C_i/D_i$;
%     \EndFor
%     \If {$D > n$}
%         \State return  false;
%     \Else
%         \State return true;
%     \EndIf

%   \end{algorithmic}
% \end{algorithm}

%%%%%%%%%%%%%%%%%%%%%%%%%%%%%%%%%%%%%%%%%%%%%%%%%%%%%%%%%%%%%%%%%%%%%%%%%%%%%%%%
\section{Experimental Evaluation}
\label{sec:exp}

We present our experiments to show that our CCL outperforms the state-of-the-art continual learning method in dealing with catastrophic forgetting for LPR methods and enables LPR methods to achieve stronger generalization ability compared to existing metric learning schemes.
We further provide ablation studies to validate each part of CCL.

\subsection{Experimental Setups}
\subsubsection{Datasets} \revise{We select three large-scale point cloud datasets to evaluate our method, including Oxford RobotCar~\cite{maddern2017ijrr}, MulRan~\cite{kim2020mulran}, and \chen{PointNetVLAD (PNV)}~\cite{angelina2018cvpr}. MulRan~\cite{kim2020mulran} consists of two separate datasets, DCC collected in an urban environment and Riverside collected on roads and bridges along a river.
The detailed dataset information and train/test setups are shown in~\tabref{tab:settings}.
We follow previous place recognition protocols to remove the ground plane of all LiDAR scans and sample the rest to 4096 points with the coordinates normalized between [-1, 1]. 
During training on Oxford~\cite{maddern2017ijrr} and PNV~\cite{angelina2018cvpr} datasets, if the distance between two scans is less than 10\,m or more than 50\,m, we treat them as a positive or negative pair respectively. During testing, if the distance is within 25\,m, we treat them as a positive pair.}
For training on MulRan~\cite{kim2020mulran}, the thresholds of positive and negative pairs are 10\,m and 20\,m. And during testing, the threshold of the positive pair is 10\,m.
 
\subsubsection{LPR methods} We choose three representative LPR methods PointNetVLAD~\cite{angelina2018cvpr}, LoGG3D-Net~\cite{vidanapathirana2022logg3d} and MinkLoc3D~\cite{komorowski2021minkloc3d} as backbones to validate our method. For LoGG3D-Net~\cite{vidanapathirana2022logg3d}, we replace the second-order pooling~\cite{vidanapathirana2021locus} with max-pooling module and drop the point-based local consistency loss since they cannot work on nonadjacent datasets. 
We use the same projector with two fully connected layers with the size of 256 for all backbones in all experiments.

\subsubsection{Evaluation protocols}
\revise{For evaluating the catastrophic forgetting, we follow InCloud~\cite{knights2022incloud} to use a 4-step protocol and apply our CCL in the following order: $\text{Oxford}\,\rightarrow\,\text{DCC}\,\rightarrow \,\text{Riverside}\,\rightarrow\,\text{PNV}$. 
Two baselines are chosen in the following experiments: the vanilla fine-tuning and InCloud. 
To ensure equitable comparison, we employ identical backbones and datasets across all methods, while employing distinct training strategies.
Training with InCloud~\cite{knights2022incloud} uses past samples for review and constructs the knowledge distillation loss, which is not applied when using fine-tuning.}
In line with InCloud~\cite{knights2022incloud}, we use the mean Recall@1 (mR@1) and Forgetting score (F) as defined below to quantitatively evaluate the capabilities of overcoming catastrophic forgetting.
\begin{align}
\vspace{-0.1cm}
    \text{mR@1}&=\frac{1}{T} \sum_{t=1}^{T}{\text{R}_{T,t}} \\
    \text{F}&=\frac{1}{T-1} \max_{l\in 1\dots t} \sum_{t=1}^{T}{\{\text{R}_{l,t}\}}-\text{R}_{T,t}
\vspace{-0.1cm}
\end{align}
where $T$ is the total number of training environments and $t$ is the current training sequence. $\text{R}_{i,j}$ means the recall@1 of test set $j$ after training step $i$.
We use a batch size of 32 and the memory size is set to 512 in line with InCloud~\cite{knights2022incloud}.
The size of the feature pool is set to 10,000 for the first training on Oxford dataset~\cite{maddern2017ijrr} and 1,000 for the rest three datasets.
We set the momentum $m = 0.99$, the contrastive temperature $\tau = 0.07$ and the distillation temperature ${\tau'} = 0.1$. All our experiments are trained on the same machine with a single Nvidia GeForce 2080 GPU for a fair comparison.
\vspace{-0.1cm}
\begin{table}[t]
    \renewcommand\arraystretch{1.1}
    \setlength{\tabcolsep}{1.8pt}
    \centering
    \caption{Datasets information}
    \begin{tabular}{l|C{1.2cm}C{1.2cm}C{0.8cm}C{0.8cm}C{1.2cm}C{1cm}}
            \toprule
            \specialrule{0em}{1.2pt}{1.2pt}
            Dataset & LiDAR Sensor & Collected City & Train Size & Test Size& Database Date & Query Date \\
            \specialrule{0em}{1.2pt}{1.2pt}
            \midrule
            \specialrule{0em}{1.2pt}{1.2pt}
            Oxford\cite{maddern2017ijrr} & SICK LMS-151 & Oxford & 22k & 3k & 05/2014 & 11/2015 \\
            DCC\cite{kim2020mulran} & Ouster OS1-64 & Dajeon & 5.5k & 15k & 08/2019 & 09/2019 \\
            Riverside\cite{kim2020mulran} & Ouster OS1-64 & Sejong & 5.5k & 18.6k & 08/2019 & 08/2019 \\
            \revise{PNV\cite{angelina2018cvpr}} & \revise{Velodyne HDL-64} & \revise{Singapore} & \revise{6.6k} & \revise{1.8k} & \revise{10/2017} & \revise{10/2017} \\
            \bottomrule
    \end{tabular}
    \label{tab:settings}
    \vspace{-0.4cm}
\end{table}

%%%%%%%%%%%%%%%%%%%%%%%%
\subsection{Evaluation on Overcoming Catastrophic Forgetting}

\begin{figure}[t]
  \centering
  \includegraphics[width=0.95\linewidth]{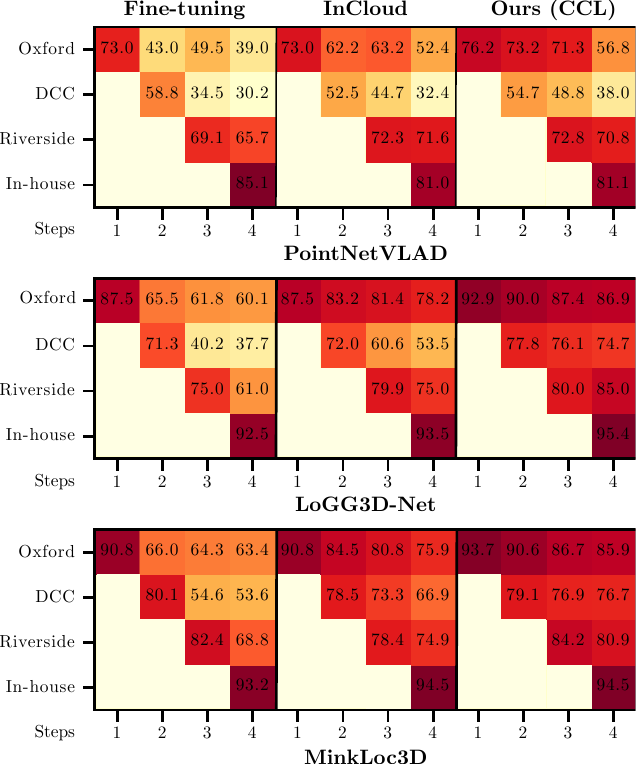}
    \caption{Average Recall@1 results using the 4-step protocol. Three sets of results from left to right for each backbone correspond to fine-tuning, InCloud, and CCL respectively.
    }
    \vspace{-0.3cm}
  \label{fig:heatmap}
\end{figure}

The first experiment supports that our CCL overcomes catastrophic forgetting between domain gaps for LPR and outperforms the state-of-the-art continual learning place recognition method InCloud~\cite{knights2022incloud}.
In every step, we train the model on the training set of the current dataset and evaluate it on all the test sets, including the current and past datasets.
As shown in~\figref{fig:heatmap}, from left to right are evaluating recall@1 results of using fine-tuning, InCloud~\cite{knights2022incloud} and our CCL respectively.
As can be seen, our method achieves the highest recall scores on all three backbones after training on the four datasets continually. Benefiting from its continual learning strategy, InCloud~\cite{knights2022incloud} maintains a higher recall than using fine-tuning only. However, it performs worse than our CCL.
Note that our CCL maintains the average recall@1 on Oxford~\cite{maddern2017ijrr} 4.4\%, 8.7\%, and 10.0\% more than InCloud~\cite{knights2022incloud} even after 4-step training.
In every step, our CCL consistently outperforms the other two methods, which shows the overall superiority of our method in overcoming catastrophic forgetting. 

\begin{table}[t]
  \caption{Continual learning results trained on the 4-step protocol.}
  \renewcommand\arraystretch{1.1}
  \setlength{\tabcolsep}{13.5pt}
  \centering
  \begin{tabular}{lc|cc}
    \toprule
    LPR backbones & Method & mR@1$\uparrow$  & F$\downarrow$ \\
    \midrule
    \multirow{3}{*}{PointNetVLAD\cite{angelina2018cvpr}} & Fine-tuning & 54.8 & 22.0 \\
    & InCloud & 60.5 & 13.8 \\
    & CCL (Ours) & \textbf{64.4} & \textbf{12.7} \\
    \hline
    \noalign{\vskip 0.1cm} 
    \multirow{3}{*}{LoGG3D-Net\cite{vidanapathirana2022logg3d}} & Fine-tuning & 65.3 & 25.0 \\
    & InCloud & 76.5 & 10.9 \\
    & CCL (Ours) & \textbf{84.6} & \textbf{1.4} \\
    \hline
    \noalign{\vskip 0.1cm} 
    \multirow{3}{*}{MinkLoc3D\cite{komorowski2021minkloc3d}} & Fine-tuning & 71.7 & 22.5 \\
    & InCloud & 79.9 & 10.0 \\
    & CCL (Ours) & \textbf{84.9} & \textbf{4.5} \\
    \bottomrule
  \end{tabular}
  \label{tab:mrecall_forgetting}
\vspace{-0.4cm}
\end{table}

To compare different methods numerically, we provide the average mR@1 and F scores across all datasets in~\tabref{tab:mrecall_forgetting}. Compared with InCloud~\cite{knights2022incloud} and fine-tuning, our method achieves the highest recall and least forgetting with all different place recognition backbones.
This validates again that our method improves the performance of different LPR methods consistently and significantly, which can be applied as a common strategy for LPR overcoming catastrophic forgetting.

%%%%%%%%%%%%%%%%%%%%%%%%
\subsection{Evaluation on Generalization Performance}

The second experiment evaluates the LPR performance and generalization of different learning schemes. We use the average recall@1 and recall@1\% as evaluation metrics and compare different methods using different datasets.
\revise{We choose Oxford~\cite{maddern2017ijrr} and three datasets from PNV~\cite{angelina2018cvpr}, namely University, Resident, and Business, evaluating our method thoroughly.}
Specifically, we train the backbones on one dataset and directly evaluate the other three datasets. 
The subscripts O, U, R, and B represent training on the Oxford, University, Resident, and Business datasets respectively.
We compare our method using contrastive learning with the triplet learning applied by InCloud~\cite{knights2022incloud}.
The results are shown in~\tabref{tab:generalization}. 
Our method improves the recall@1 and recall@1\% significantly and outperforms the baseline method on every set of experiments by a large margin.
The size of the University, Resident, or Business training set is about 10\% of Oxford. 
Although only trained on a small dataset, our model achieves considerable recall on Oxford.
Such results show that using our method, the LPR backbones can learn a more transferable representation thus generalizing well into different environments.

\begin{table*}[t]
    \renewcommand\arraystretch{1.1}
    \setlength{\tabcolsep}{10pt}
    \centering
        \caption{ Evaluation of Recall on Multi Datasets$^*$}
        \begin{tabular}{l|cccccccc}
            \toprule
            \multirow{2}{*}{LPR Backbones} & \multicolumn{2}{c}{Oxford} & \multicolumn{2}{c}{University}  & \multicolumn{2}{c}{Resident}  & \multicolumn{2}{c}{Business}\\
            \cmidrule(lr){2-3} \cmidrule(lr){4-5} \cmidrule(lr){6-7} \cmidrule(lr){8-9} & Original & Ours & Original & Ours & Original & Ours & Original & Ours \\
            \midrule
            $\text{PointNetVLAD}_\text{O}$\cite{angelina2018cvpr} & 73.0\,/\,86.6 & \textbf{80.2\,/\,90.4} & 65.6\,/\,78.3 & \textbf{77.8\,/\,88.8} & 58.9\,/\,73.8 & \textbf{71.5\,/\,80.9} & 60.4\,/\,67.4 & \textbf{69.3\,/\,75.9} \\
            $\text{PointNetVLAD}_\text{U}$\cite{angelina2018cvpr} & 34.1\,/\,49.9 & \textbf{53.1\,/\,67.5} & 85.2\,/\,94.3 & \textbf{91.0\,/\,97.4} & 80.4\,/\,89.4 & \textbf{83.8\,/\,92.1} & 77.7\,/\,84.6 & \textbf{81.0\,/\,86.5} \\
            $\text{PointNetVLAD}_\text{R}$\cite{angelina2018cvpr} & 40.5\,/\,56.2 & \textbf{47.3\,/\,62.0} & 82.5\,/\,92.3 & \textbf{87.9\,/\,95.8} & 84.6\,/\,93.2 & \textbf{87.2\,/\,94.5} & 79.1\,/\,85.0 & \textbf{80.8\,/\,86.6} \\
            $\text{PointNetVLAD}_\text{B}$\cite{angelina2018cvpr} & 45.1\,/\,60.6 & \textbf{48.4\,/\,63.4} & 82.1\,/\,93.0 & \textbf{87.3\,/\,94.8} & 81.0\,/\,89.7 & \textbf{86.2\,/\,92.8} & 83.3\,/\,89.6 & \textbf{86.7\,/\,91.4} \\
            $\text{LoGG3D-Net}_\text{O}$\cite{vidanapathirana2022logg3d} & 87.5\,/\,95.9 & \textbf{92.9\,/\,97.5} & 68.6\,/\,82.9 & \textbf{78.5\,/\,90.4} & 66.2\,/\,81.7 & \textbf{76.1\,/\,87.0} & 71.2\,/\,79.6 & \textbf{78.1\,/\,84.9} \\
            $\text{LoGG3D-Net}_\text{U}$\cite{vidanapathirana2022logg3d} & 56.2\,/\,73.3 & \textbf{62.0\,/\,76.6} & 91.8\,/\,98.5 & \textbf{96.2\,/\,98.9} & 86.6\,/\,95.4 & \textbf{93.5\,/\,97.7} & 86.0\,/\,92.1 & \textbf{91.6\,/\,95.6} \\
            $\text{LoGG3D-Net}_\text{R}$\cite{vidanapathirana2022logg3d} & 59.3\,/\,75.9 & \textbf{64.0\,/\,77.5} & 87.6\,/\,95.9 & \textbf{92.7\,/\,98.7} & 91.9\,/\,97.5 & \textbf{95.6\,/\,99.2} & 88.5\,/\,93.1 & \textbf{91.3\,/\,94.4} \\
            $\text{LoGG3D-Net}_\text{B}$\cite{vidanapathirana2022logg3d} & 53.4\,/\,70.2 & \textbf{62.8\,/\,77.0} & 83.1\,/\,94.1 & \textbf{91.3\,/\,97.9} & 83.7\,/\,93.3 & \textbf{91.6\,/\,95.9} & 95.5\,/\,98.8 & \textbf{98.9\,/\,99.8} \\
            $\text{MinkLoc3D}_\text{O}$\cite{komorowski2021minkloc3d} & 90.8\,/\,97.0 & \textbf{93.7\,/\,97.4} & 81.6\,/\,92.2 & \textbf{87.2\,/\,95.0} & 80.8\,/\,89.7 & \textbf{87.4\,/\,93.2} & 77.3\,/\,84.0 & \textbf{84.9\,/\,89.8} \\
            $\text{MinkLoc3D}_\text{U}$\cite{komorowski2021minkloc3d} & 58.4\,/\,73.9 & \textbf{67.5\,/\,80.8} & 93.0\,/\,98.6 & \textbf{96.8\,/\,99.4} & 86.0\,/\,94.1 & \textbf{93.8\,/\,97.1} & 86.3\,/\,92.1 & \textbf{92.8\,/\,96.0} \\
            $\text{MinkLoc3D}_\text{R}$\cite{komorowski2021minkloc3d} & 57.6\,/\,73.7 & \textbf{66.2\,/\,79.7} & 89.4\,/\,96.1 & \textbf{94.0\,/\,98.7} & 89.5\,/\,96.4 & \textbf{93.7\,/\,97.4} & 86.4\,/\,91.7 & \textbf{92.0\,/\,95.1} \\
            $\text{MinkLoc3D}_\text{B}$\cite{komorowski2021minkloc3d} & 55.7\,/\,72.1 & \textbf{61.2\,/\,76.2} & 86.8\,/\,95.7 & \textbf{91.5\,/\,97.6} & 83.5\,/\,92.4 & \textbf{91.7\,/\,96.4} & 93.9\,/\,97.0 & \textbf{96.7\,/\,98.6} \\
            \bottomrule        
            \multicolumn{9}{p{0.95\linewidth}}{\footnotesize
                $^*:$ Average recall@1 and recall@1\% results on different datasets and methods. All methods have been trained on one of the four datasets and directly evaluated on the others. The subscript represents the training dataset as Oxford(O), University(U), Resident(R), and Business(B). The best results of the experiment are shown in bold.} 
        \end{tabular}
        \vspace{-0.4cm}
        \label{tab:generalization}
\end{table*}

\subsection{Ablation Study and Parameter Analysis}

\begin{table}[t]
    \renewcommand\arraystretch{1.1}
    \setlength{\tabcolsep}{9.6pt}
    \centering
    \caption{Ablation Study}
    \begin{tabular}{lccc|cc}
        \toprule
            & Contrastive & Projector & KD & mR@1$\uparrow$ & F$\downarrow$  \\
        \midrule
        $[\text{a}]$ & $\times$ & $\times$ & $\times$ & 63.9 & 23.2 \\
        $[\text{b}]$ & $\surd$ & $\times$ & $\times$ & 67.0 & 22.9 \\
        $[\text{c}]$ & $\surd$ & $\surd$ & $\times$ & 72.7 & 15.2 \\
        $[\text{d}]$ & $\surd$ & $\surd$ & $all$ & 66.9 & 16.0 \\
        $[\text{e}]$ & $\surd$ & $\surd$ & $\surd$ & \textbf{78.0} & \textbf{6.2} \\
        \bottomrule
    \end{tabular} \\
    \label{tab:ablation}
    \vspace{-0.2cm}
\end{table}
In this section, we further provide studies to validate the effectiveness of each component of our CCL and support our parameter choosing. 

The ablation study of different components of our CCL is shown in~\tabref{tab:ablation}. 
\revise{The results are the mR@1 and F scores averaged over three place recognition backbones, including PointNetVLAD~\cite{angelina2018cvpr}, LoGG3D-Net~\cite{vidanapathirana2022logg3d} and MinkLoc3D~\cite{komorowski2021minkloc3d} after training using the 4-step protocol on Oxford~\cite{maddern2017ijrr}, MulRan~\cite{kim2020mulran} and PNV~\cite{angelina2018cvpr} datasets.} 
Therefore, such results are statistically significant and can reflect the general effectiveness of our method.
As can be seen in~\tabref{tab:ablation},
[a] shows the results of vanilla fine-tuning via training with the 4-step protocol.
[b] represents the results of using contrastive loss without the projector trained with the 4-step protocol.
[c] uses contrastive loss as well as the projector during training.
[d] includes the memory review and the knowledge distillation strategy proposed in InCloud~\cite{knights2022incloud} and marked as "all". This experiment is to show our devised contrastive memory bank fits CCL and tackles forgetting better.
[e] shows the results of using the full pipeline of the proposed CCL. 
The overall ablation study shows that every component in our CCL positively impacts the LPR performance, and our CCL achieves the best performance when combining all the proposed modules. 

We provide analyses of the key parameters of our method in \tabref{tab:feature_size}. 
We use MinkLoc3D~\cite{komorowski2021minkloc3d} backbone to search parameters on the Oxford~\cite{maddern2017ijrr} dataset and fix them in all experiments.
The upper part shows the results with respect to different feature pool sizes, and the best result is achieved using 10,000 negatives.
In the low part, we test different momentum sizes using 10,000 negatives, and the best $m$ is 0.99.

\begin{table}[t]
    \renewcommand\arraystretch{1.1}
    \setlength{\tabcolsep}{5.5pt}
    \centering
    \caption{Parameter Analysis}
    \begin{tabular}{l|c|cccc}
            \toprule
            Parameters & Values & Oxford & University & Resident & Business \\
            \midrule
            \multirow{5}{1.6cm}{Feature Pool Size} &100 & 91.4 & 85.1 & 81.4 & 82.7 \\
            &1,000 & 93.5 & 88.0 & 86.0 & 83.4 \\
            &5,000 & 93.7 & 87.0 & 86.0 & 83.6 \\
            &10,000 & \textbf{93.7} & \textbf{87.2} & \textbf{87.4} & \textbf{84.9} \\
            &15,000 & 93.3 & 85.5 & 83.5 & 82.6 \\    
            \midrule
            \multirow{3}{1.5cm}{Momentum size} &0.9 & 93.1 & 85.5 & 83.2 & 82.6 \\
            &0.99 & \textbf{93.7} & \textbf{87.2} & \textbf{87.4} & \textbf{84.9} \\
            &0.999 & 93.2 & 86.8 & 85.3 & 82.8 \\
            \bottomrule
    \end{tabular} \\
    \label{tab:feature_size}
    \vspace{-0.4cm}
\end{table}

% \begin{table}[t]
%     \renewcommand\arraystretch{1.2}
%     \setlength{\tabcolsep}{1pt}
%     \centering
%     \caption{Momentum size}
%         \begin{tabular}{l|cccc}
%             \toprule
%             m & Oxford & University & Resident & Business \\
%             \midrule
%             0.9 & 93.1 & 85.5 & 83.2 & 82.6 \\
%             0.99 & \textbf{93.7} & \textbf{87.2} & \textbf{87.4} & \textbf{84.9} \\
%             0.999 & 93.2 & 86.8 & 85.3 & 82.8 \\
%             \bottomrule
%         \end{tabular} \\
%     \label{tab:momentum}
% \end{table}

%%%%%%%%%%%%%%%%%%%%%%%%
\subsection{Discussion and Insights}

\begin{figure*}[t]
  \centering
  \includegraphics[width=0.88\linewidth]{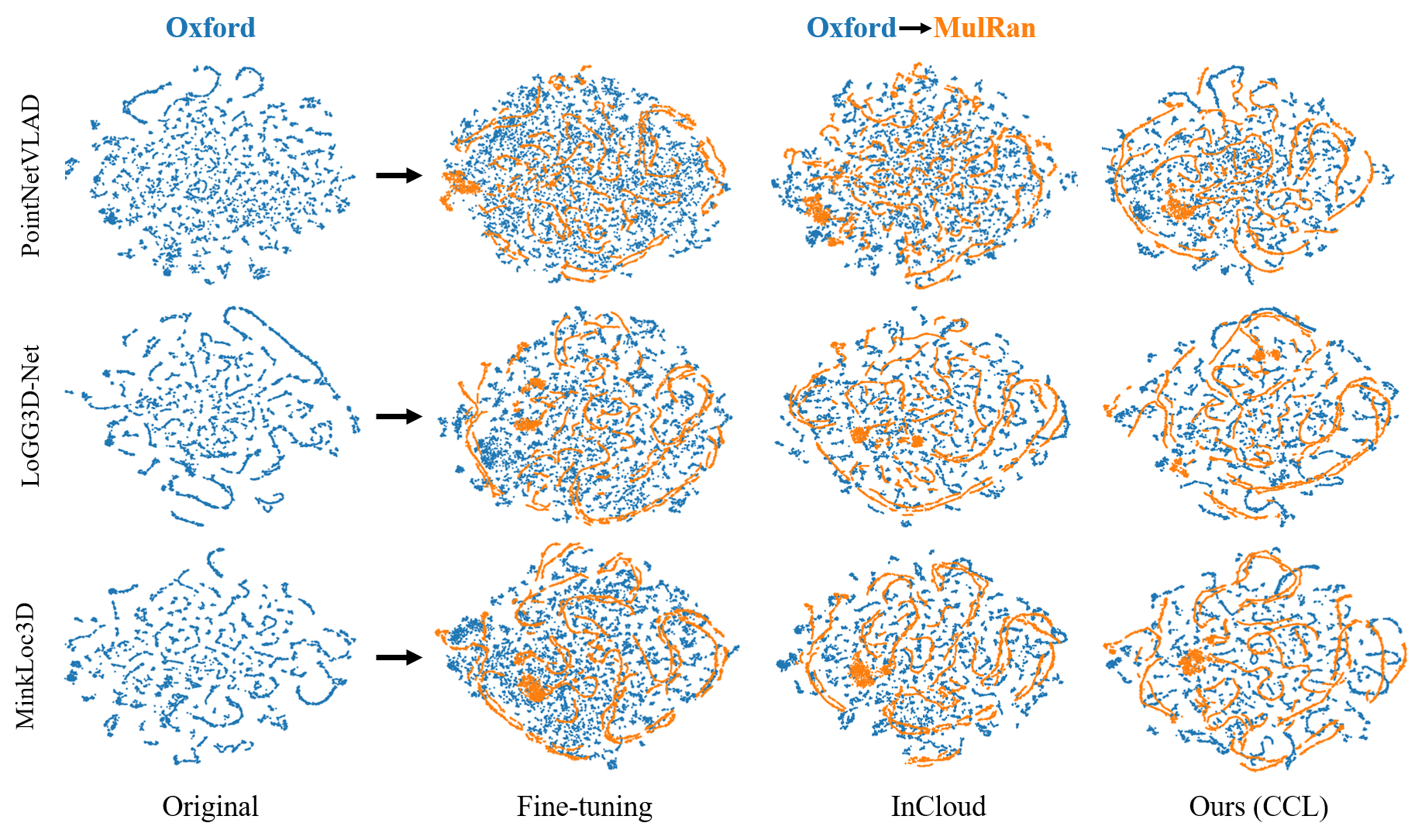}
  \caption{t-SNE visualization of LPR descriptors. The first column is the feature distribution in the embedding space generated by the original LPR methods trained on Oxford. The rest three columns are the combined feature distribution from two environments generated by the updated network models using different continual learning methods.}
  \label{fig:tsne}
  \vspace{-0.5cm}
\end{figure*}

\revise{To provide more insights, following InCloud~\cite{knights2022incloud}, we utilize t-SNE~\cite{van2008visualizing} to visualize the changes in global descriptor distribution in the embedding space.
The model is first trained on Oxford \cite{maddern2017ijrr} (blue) and then trained on DCC dataset~\cite{kim2020mulran} (orange).}
As shown in~\figref{fig:tsne}, the original LPR descriptors trained on Oxford~\cite{maddern2017ijrr} are gathered separately.
After training under the fine-tuning strategy, the intricate structure of the embedding space will inevitably result in a collapse.
InCloud~\cite{knights2022incloud} applying a continual learning strategy can alleviate the collapse, but the features are less descriptive, as shown by shorter and noisier feature clusters. 
Our method under the proposed contrastive and continual combined learning %via natural memory review and the feature distribution-based knowledge distillation 
preserves most of the structure of the embedding space between different environments. This visualization further provides insights into our method of generating more transferrable contrastive features and overcoming catastrophic forgetting.
Although using thousands of negatives, the training time and memory consumption are almost unchanged compared to the original LPR methods. 
% In summary, our evaluation suggests that our CCL achieves state-of-the-art continual contrastive learning performance. 
% Our method achieves strong generalization ability and overcomes catastrophic forgetting.
% Thus, we support all our claims with this experimental evaluation.

%%%%%%%%%%%%%%%%%%%%%%%%%%%%%%%%%%%%%%%%%%%%%%%%%%%%%%%%%%%%%%%%%%%%%%%%%%%%%%%%
\section{Conclusion}
\label{sec:conclusion}

In this paper, we presented a novel approach combining contrastive and continual learning for the LPR task.
Our method constructs a large negative feature pool and trains the LPR methods using contrastive loss, which enables LPR methods to generate more descriptive and generalized descriptors. Based on them, our method conducts feature distribution-based knowledge distillation loss for past samples to successfully overcome the LPR catastrophic forgetting problem.
We implemented and evaluated our approach on multiple datasets in different environments
and provided comparisons to the existing state-of-the-art continual learning method. Our experiments suggest that by combining contrastive and continual learning, our method significantly outperformed the baseline method in tackling catastrophic forgetting and consistently improved the generalization ability of learning-based LPR methods.
% \todo{Future directions: The training sets are now constructed according to the ground truth position which is sometimes not available. Also, the principal of defining positives and negatives makes no sense in some specific conditions. Negatives in one setting may be positives in another. All positives are not explicitly differentiated. Therefore, unsupervised contrastive continual learning for place recognition is worth exploring.}

%%%%%%%%%%%%%%%%%%%%%%%%%%%%%%%%%%%%%%%%%%%%%%%%%%%%%%%%%%%%%%%%%%%%%%%%%%%%%%%%
%% Future work: Use only if applicable -- but if so, use the following
%% sentence to start:
% Despite these encouraging results, there is further space for improvements. 

%%%%%%%%%%%%%%%%%%%%%%%%%%%%%%%%%%%%%%%%%%%%%%%%%%%%%%%%%%%%%%%%%%%%%%%%%%%%%%%%
% Only if applicable
%\section*{Acknowledgments}
%We thank XXX for fruitful discussions and for \dots

% \bibliographystyle{plain_abbrv}
\bibliographystyle{IEEEtran}

% All new citations should go to new.bib. The file glorified.bib should go
% be the one from the ipb server. After paper or related work has been
% written merge the entries from new.bib to glorified.bib ON THE SERVER,
% replace the glorified.bib in this repository and empty the new.bib
\bibliography{glorified,new}

\end{document}